\documentclass[sigconf]{acmart}
\acmConference[KDF '23]{The 4th Workshop on Knowledge Discovery from Unstructured Data in Financial Services}{July 27,
  2023}{Taipei, Taiwan}
\acmDOI{XXXXXXX.XXXXXXX}
\usepackage{multirow} 
\usepackage{float}
\usepackage{xcolor}

\begin{document}

\title{iMETRE: Incorporating Markers of Entity Types for Relation Extraction}

\author{N Harsha Vardhan}
\affiliation{%
  \institution{International Institute of Information Technology}
  \city{Hyderabad}
  \country{India}
}
\email{nemani.v@research.iiit.ac.in}

\author{Manav Chaudhary}
\affiliation{%
  \institution{International Institute of Information Technology}
  \city{Hyderabad}
  \country{India}
}
\email{manav.chaudhary@research.iiit.ac.in}

\begin{abstract}
Sentence-level relation extraction (RE) aims to identify the relationship between 2 entities given a contextual sentence. While there have been many attempts to solve this problem, the current solutions have a lot of room to improve. In this paper, we approach the task of relationship extraction in the financial dataset REFinD \cite{kaur2023refind}. Our approach incorporates typed entity markers representations, and various models finetuned on the dataset, which has allowed us to achieve an ${F_1}$ score of 69.65\% on the validation set. Through this paper, we discuss various approaches and possible limitations.

\end{abstract}


\maketitle

\section{Introduction}
To extract meaningful semantic relationships from textual data, in the field of natural language processing (NLP) and information retrieval, relation extraction (RE) plays a defining role. The task involves the categorization of the connections between two entities of specific types, such as persons, organizations, or locations, into various semantic categories like financial relationships, employment associations, or geographical affiliations, among others.

Although large-scale datasets produced from generic knowledge sources like Wikipedia, online texts, and news articles have significantly improved the performance of existing models for relation extraction, these resources frequently fall short in capturing domain-specific hurdles. Financial text documents (such as financial reports and Securities and Exchange Commission (SEC) filings) require complicated extraction methods, presenting a unique set of difficulties that necessitate domain-specific extraction methods. These documents have entities and relations that involve but are not limited to numbers, money, dates, legal information, and claims. Additionally, the sentences in such financial documents are lengthier, more complex, and cover a broader range of entity distances.


\begin{figure}
  \centering
  \includegraphics[width=0.49\textwidth]{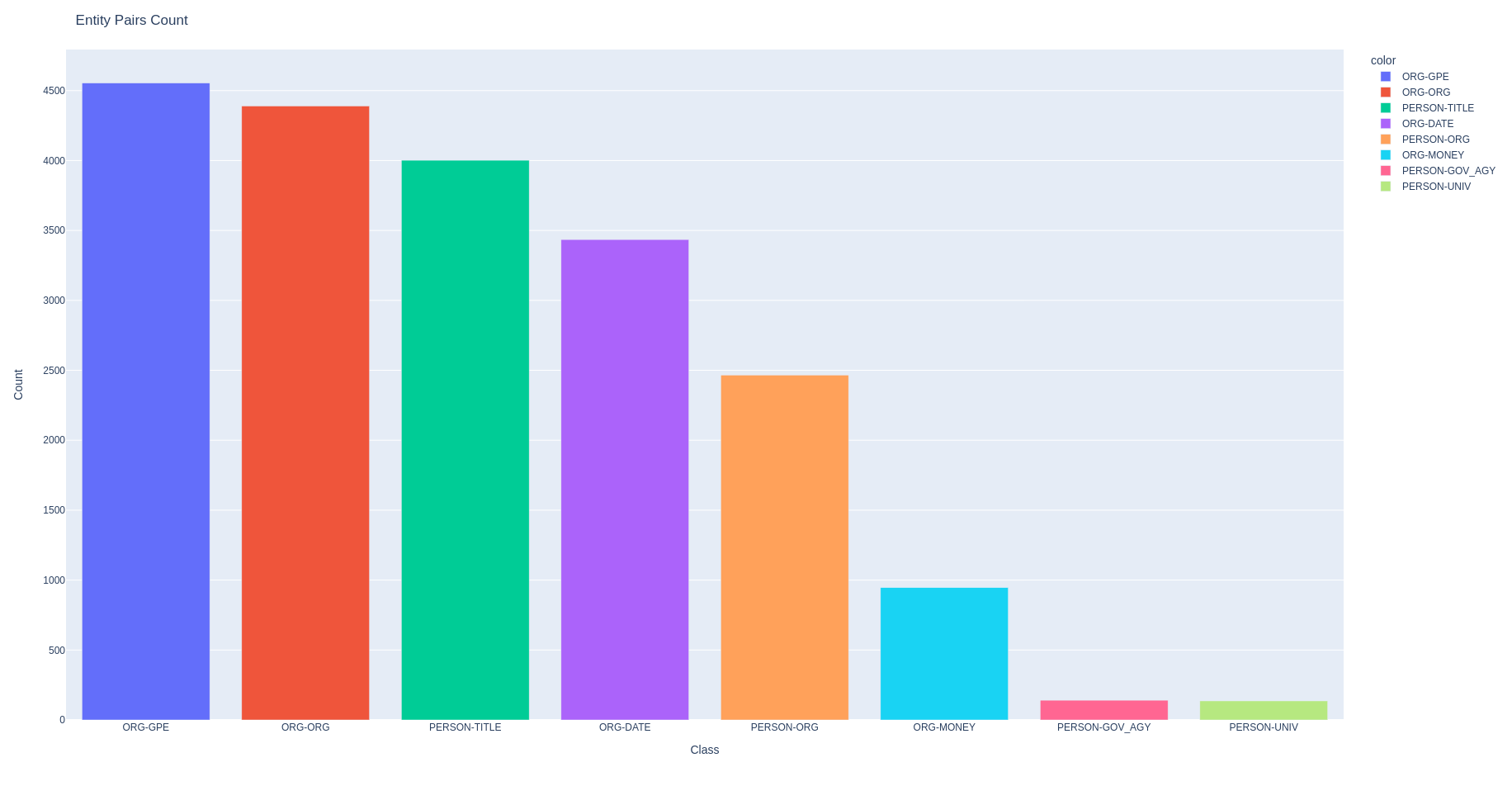}
  \caption{Count of Entity Pairs in the ReFind Dataset}
  \label{fig:EntityPairs}
\end{figure}

The REFinD dataset\cite{kaur2023refind} is the largest relation extraction dataset for financial documents to date and contains ~29K instances and 22 relations among 8 types of entity pairs. REFinD is a domain-specific financial relation extraction dataset created using raw text from various 10-X reports (including 10-K, 10-Q, etc., broadly known as 10-X) of publicly traded companies obtained from US Securities and Exchange Commission (SEC) which is a rich and complex data source.

Compared to other well-known datasets for relation extraction such as TACRED (with 32 relations and a majority (79.5\%) of which being NO\_RELATION instances), REFinD is a much more balanced dataset with only 45.5\% being NO\_RELATION and has a higher number of instances for each relation it covers. Among financial datasets, both FinRED \cite{sharma2022finred} and CorpusFR are smaller compared to REFinD and have fewer relation types. REFinD also contains longer sentences with an average sentence length of 53 words and average contextual complexity of 11 words between entity pairs.

\begin{figure}
  \centering
  \includegraphics[width=0.45\textwidth]{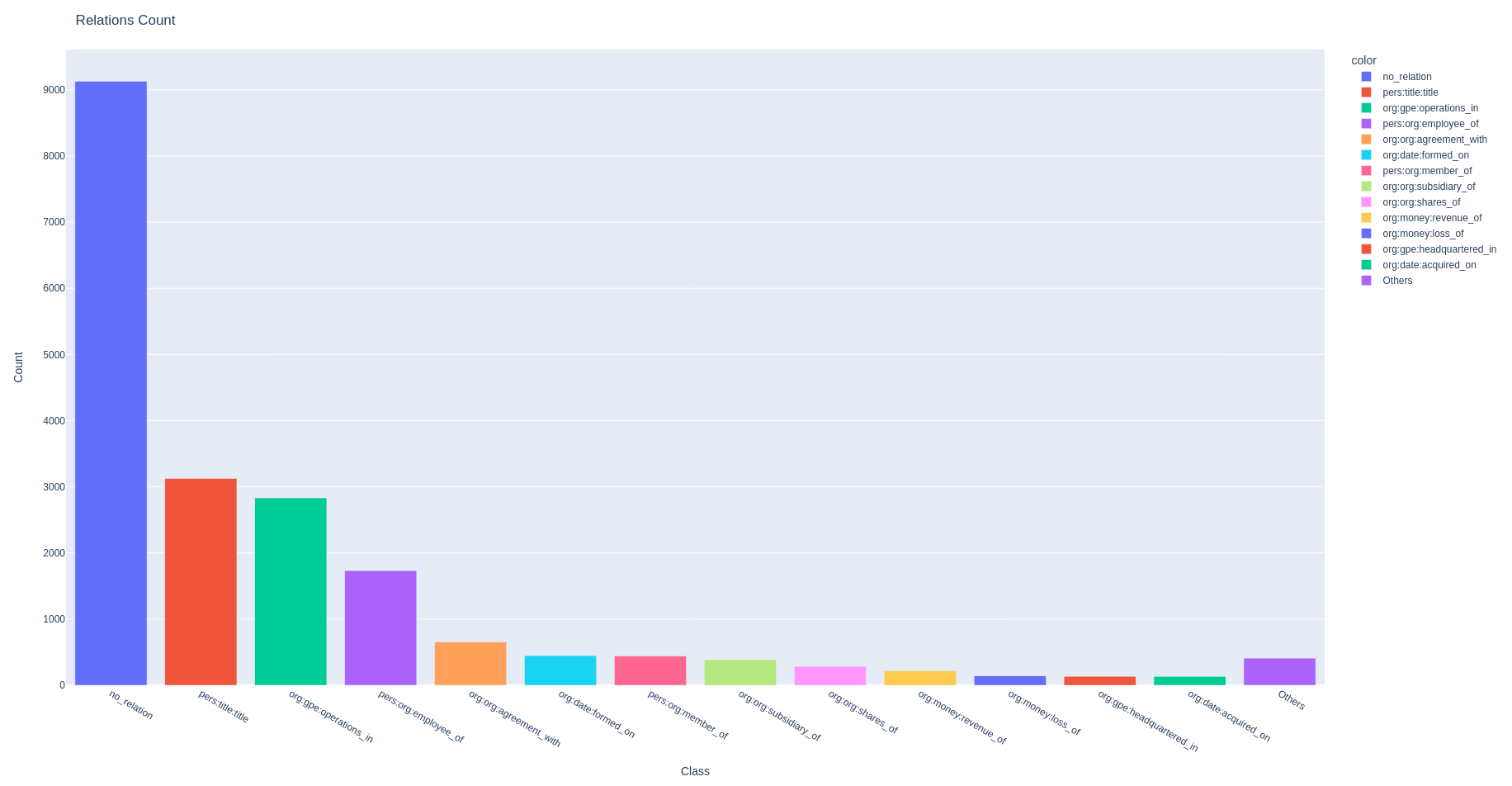}
  \caption{Count of Relations in the ReFind Dataset}
  \label{fig:Relations}
\end{figure}

This research endeavor aims to bridge the gap between general-purpose relation extraction datasets and the unique demands of the finance domain. By utilizing REFinD, researchers and practitioners in the field of financial NLP can access a rich and extensive resource that facilitates the extraction of financial entities and their intricate relationships from the text. This dataset holds great potential for advancing various tasks, including constructing knowledge graphs, question-answering systems, and semantic search engines tailored to the financial domain.

The main contributions of this paper include the following:
\begin{itemize}
\item \textbf{Markers for Entity Types:} This approach incorporates both entity types and spans into the context for classification, leading to accurate relationship predictions. Typed Entity Markers are used in PLM-based RE models to mark entity spans and types using punctuation \cite{zhou2021improved}, enhancing contextual embeddings and improving the model's understanding of entity relationships. 
\end{itemize}
\begin{itemize}
\item \textbf{Model Exploration:} We explore and compare the performance of various transformer-based models, including BERT-Base\cite{devlin2018bert}, RoBERTa-Base\cite{liu2019roberta}, DistilBERT\cite{sanh2019distilbert}, LUKE-Base\cite{yamada2020luke}, XLNet-Base\cite{yang2019xlnet}, and FLANG-DistilBERT\cite{shah2022flue}, in the context of sentence-level relation extraction and the ReFinD dataset, which provides valuable insights into the effectiveness of different models in capturing entity relationships in the financial domain.
\end{itemize}
\begin{itemize}
\item \textbf{Entity Type Pair Classification:} To address the weak influence of entity types and the ambiguity between semantically close relationships, we discuss a novel approach of dividing the classification task into eight tasks based on entity type pairs. This approach leads to improved classification accuracy and reduces error rates by considering the specific characteristics of each entity type pair.
\end{itemize}



\section{Methodology}

In this section, we first formally define the problem in the context of the ReFinD dataset while also presenting various model architectures and experiments performed along with the quantified metrics.
\subsection{Problem Definition}
In this task, we focus on sentence-level Relation Extraction. Specifically , given a sentence $x$ and an entity pair $(e_{1},e_{2})$, the objective is to predict the relationship $r$ between $e_1$ and $e_2$ in the sentence $x$. 

\subsection{Model Architecture}
The Relation Extraction(RE) classifier is based on the Permutated Language Modelling (PLM) objective-based RE models. Given the input sentence $x$, we mark the entity spans and entity types using a variation of the typed entity marker\cite{soares2019matching}. To this end, we mark an entity type and span using punctuation marks \cite{zhou2021improved}. Specifically, the subject entity is enclosed with \textcolor{olive}{" @ "} and the object entity with \textcolor{red}{" \# "}. The entity types are also represented using the label text, which is prepended to the entity spans and is enclosed by \textcolor{orange}{" * "} for subjects and \textcolor{cyan}{" $\wedge$ "} for objects. For instance, given a text, the modified text would be  ..."\textcolor{olive}{@ \textcolor{orange}{*subj-type*} SUBJ @} $\ldots$ \textcolor{red}{$\#$ \textcolor{cyan}{$\wedge$ obj-type $\wedge$} OBJ $\#$}  ".... This allows us to include both; the type of entities, as well as the entity spans into context for classification. This is a variation of many such similar techniques, entity masks, and markers which are well-defined solutions in the task of RE.

We then feed the generated sentence to a model to get contextual embeddings and modify the model's last layer to a softmax classifier to obtain inferences. In inference, the class with the highest softmax probability is predicted as the relationship. We use a batch size of 8 and finetune the model for the training dataset for five epochs. Also, the reported metric on the test set is $F_1$, while on the validation set is a stricter $ F_1$ metric. More specifically, while calculating $F_1$, we remove all instances where the model correctly predicts NO\_RELATION and use the remaining instances to calculate $F_1$.

\subsection{Experiments}
In this section, we evaluate and report proposed techniques on the REFinD dataset. We primarily tried two approaches: one, we switch the models while keeping the configurations constant for better metrics, and two, we individually try to model pair-pair relation extraction accurately and use multiple models for inference which depend on the involved entity type pair.

\subsubsection{Switching Architectures}
Using the above data processing methods for capturing entity spans and types, we employ various models, namely BERT-Base, RoBERTa-Base, DistilBert, LUKE-Base, XLNET-Base, and FLANG-DistilBERT, for capturing entity spans and types, each of which is trained on different datasets and subsequently fine-tuned. The results obtained from these models have been compiled and are presented in \hyperref[tab:table1]{Table 1}.

\begin{itemize}
    \item BERT-Base\cite{devlin2018bert} : BERT is a transformer-based model which utilizes bidirectional context to generate word representations. BERT-Base is trained on general text like English Wikipedia and books.
    \item RoBERTa-Base\cite{liu2019roberta} : Robustly Optimized BERT is an optimized variant of BERT that achieves improved performance by utilizing better training procedures, dynamic masking, and pretraining on a larger corpus.
    \item DistilBERT\cite{sanh2019distilbert} : DistilBERT is a condensed version of BERT that retains most of its performance while reducing the model size and computational requirements using the concept of knowledge distillation, where a smaller model is trained to mimic the behavior of the larger BERT model, resulting in a compact yet efficient representation.
    \item LUKE-Base\cite{yamada2020luke} : LUKE is a new pre-trained contextualized representation of words and entities based on transformer.LUKE-base utilizes a hybrid architecture combining transformer-based contextualized and knowledge-based embeddings, resulting in improved performance on entity-related tasks.
    \item XLNet-Base\cite{yang2019xlnet} : XLNET is a novel generalized permutation-based training objective that helps the model to consider all possible permutations of input during training, allowing the model to capture both bidirectional contexts which helps in differentiating between Directional Ambiguity between entities.
    \item FLANG-DistilBERT\cite{shah2022flue} :  FLANG-DistilBERT is built by further training the DistilBERT language model in the finance domain with improved performance over previous models due to the use of domain knowledge and vocabulary.
\end{itemize}

\begin{table}[H]
  \caption{Scores across Models}
  \label{tab:table1}
  \begin{tabular}{|c|c|c|}
    \hline
    \multirow{2}{*}{Model} & \multicolumn{2}{c|}{Test Set}\\
    \cline{2-3}
     & Micro $F_1$ Score & Accuracy\\
    \hline
    BERT-Base & 0.73 & 0.77\\
    \hline
    RoBERTa-Base & 0.74 & 0.75\\
    \hline
    DistilBERT & 0.73 & 0.78\\
    \hline
    XLNET-Base & 0.75 & 0.79\\
    \hline
    FLANG-DistilBERT & 0.75 & 0.79\\
    \hline
    LUKE-Base & 0.72 & 0.76\\
    \hline
  \end{tabular}
\end{table}

Since each of the models has been trained on different data and was different in terms of architecture, there have been slight increments in metrics across models.

\subsubsection{Classification based on Entity Type Pair}
One of the shortcomings of treating the task as a 22-label classification problem is that the classification is weakly affected by the entity types. This is coupled with the fact that some of the relationships are semantically close and ambiguous. For example, $member\_of$, $employee\_of$, and $founder\_of$ for entity pairs belonging to the PER-ORG, PER-UNIV, and PER-GOV groups are often confused. Since the data distribution is also non-uniform across classes, this increases the error rate. Hence we divided the classification task into eight classification tasks, one for each entity type pair, keeping a constant model DistilBERT for fine-tuning. The results obtained have been compiled and presented in \hyperref[tab:table2]{Table 2}.

\begin{table}[H]
  \caption{Cumulative Accuracy, $F_1$ Score, and Additional Metric for each Entity Pair using DistilBERT}
  \label{tab:table2}
  \begin{tabular}{|c|c|c|c|}
    \hline
    \multirow{2}{*}{Entity Pair} & \multicolumn{2}{c|}{Test Set} & \multirow{2}{*}{REFinD $F_1$ Baseline}\\
    \cline{2-3}
     & $F_1$ Score & Accuracy & \\
    \hline
    ORG-GPE & 0.79 & 0.76 & 0.85\\
    \hline
    ORG-ORG & 0.69 & 0.69 & 0.41\\
    \hline
    PERS-TITLE & 0.88 & 0.82 & 0.90\\
    \hline
    ORG-DATE & 0.91 & 0.91 & 0.81\\
    \hline
    PERS-ORG & 0.70 & 0.71 & 0.67\\
    \hline
    ORG-MONEY & 0.87 & 0.87 & 0.78\\
    \hline
    PERS-UNIV & 0.65 & 0.62 & 0.61\\
    \hline
    PERS-GOV\_AGY & 0.72 & 0.66 & 0.22\\
    \hline
  \end{tabular}
\end{table}

Although we can see better classification metrics in some entity pair classes, the overall $F_{1}$ score is relatively low. This is primarily due to the unique operationalization of the $F_{1}$ used for grading on the validation set as well as the disparity along entity-pair classes. The performance also increases in entity pair classes since some relationships are semantically similar, and even the usage of entity markers fails to compensate adequately, while segregation of such classes shows classwise better results. 

Apart from these, given the unique operationalization of the $F_{1}$ score used for grading in the shared task, approaches for an architecture that leverages the high \% of $NO\_RELATION$ entries, improvements with larger models of the architectures above mentioned can be some potential ideas of interest.

\section{Conclusion}
We see that using a modification of the classical entity mask and marker approaches which are slightly flawed due to a lack of understanding of numerical inferences (using entity markers (quote MTB)), while the proposed punctuated entity markers are better at capturing context within relations of entities as well as detecting the entity spans.

We also observe that although the results seem consistent and equitable across models, XLNET-base outperforms the others. This can be mainly attributed to the bidirectional nature of training data which, in essence, captures the importance of sequencing between entities for relation prediction, and due to structured textual forms in REFinD dataset, permutation-based attention masking might have also played a crucial role. This also achieves an $F_1$ score of $69.65\%$ on the validation dataset.

Also, from the eight-class approach, we can see an individual increase in some classes to a high degree. This could be attributed to better separation between semantically similar results. Although this approach still fails to outperform overall $F_1$, across models, mainly due to disparity in support of the classes, it still might be better for pair-specific classifications. 

\section{Future Work}
Theoretically, using Larger Models instead of Distilled and Base Models should improve the results. Also, techniques like data augmentation to inflate classes with low support might also increase the overall efficiency and viability of the models and dataset. Some other approaches of differentiating based on entity type pairs and integrating into the final layer might also yield some interesting results.
\bibliographystyle{ACM-Reference-Format}
\bibliography{References}

\appendix
\section{Performance of Models}
\begin{table}[H]
  \caption{$F_1$ Scores across Models and Relations: Part 1}
  \label{tab:example}
  \small
  \footnotesize 
  \begin{tabular}{|c|c|c|c|}
    \hline
    \multirow{2}{*}{Class} & \multicolumn{3}{c|}{Performance on the Test Set} \\
    \cline{2-4}
     & BERT-Base & RoBERTa-Base & DistilBERT \\
    \hline
    Class 0 & 0.80 & 0.78 & 0.81 \\
    \hline
    Class 1 & 0.86 & 0.83 & 0.87 \\
    \hline
    Class 2 & 0.85 & 0.85 & 0.85 \\
    \hline
    Class 3 & 0.15 & 0.14 & 0.13 \\
    \hline
    Class 4 & 0.78 & 0.74 & 0.79 \\
    \hline
    Class 5 & 0.93 & 0.67 & 0.77 \\
    \hline
    Class 6 & 0.00 & 0.00 & 0.11 \\
    \hline
    Class 7 & 0.84 & 0.00 & 0.83 \\
    \hline
    Class 8 & 0.71 & 0.55 & 0.74 \\
    \hline
    Class 9 & 0.56 & 0.63 & 0.73 \\
    \hline
    Class 10 & 0.46 & 0.44 & 0.46 \\
    \hline
    Class 11 & 0.92 & 0.00 & 0.93 \\
    \hline
    Class 12 & 0.46 & 0.00 & 0.49 \\
    \hline
    Class 13 & 0.67 & 0.00 & 0.73 \\
    \hline
    Class 14 & 0.57 & 0.00 & 0.00 \\
    \hline
    Class 15 & 0.49 & 0.59 & 0.49 \\
    \hline
    Class 16 & 0.25 & 0.21 & 0.25 \\
    \hline
    Class 17 & 0.45 & 0.15 & 0.50 \\
    \hline
    Class 18 & 0.90 & 0.91 & 0.90 \\
    \hline
    Class 19 & 0.71 & 0.52 & 0.65  \\
    \hline
    Class 20 & 0.39 & 0.38 & 0.36 \\
    \hline
    Class 21 & 0.62 & 0.00 & 0.55 \\
    \hline
  \end{tabular}
\end{table}

\begin{table}[H]
  \caption{$F_1$ Scores across Models and Relations: Part 2}
  \label{tab:example}
  \small
  \footnotesize 
  \begin{tabular}{|c|c|c|c|}
    \hline
    \multirow{2}{*}{Class} & \multicolumn{3}{c|}{Performance on the Test Set} \\
    \cline{2-4}
     & XLNet-Base & FLANG-DistilBERT & LUKE-Base \\
    \hline
    Class 0 & 0.81 & 0.82 & 0.79 \\
    \hline
    Class 1 & 0.90 & 0.86 & 0.79 \\
    \hline
    Class 2 & 0.86 & 0.86 & 0.86 \\
    \hline
    Class 3 & 0.20 & 0.20 & 0.17 \\
    \hline
    Class 4 & 0.80 & 0.79 & 0.76 \\
    \hline
    Class 5 & 0.88 & 0.86 & 0.86 \\
    \hline
    Class 6 & 0.18 & 0.10 & 0.00 \\
    \hline
    Class 7 & 0.87 & 0.87 & 0.72 \\
    \hline
    Class 8 & 0.81 & 0.74 & 0.69 \\
    \hline
    Class 9 & 0.76 & 0.76 & 0.33 \\
    \hline
    Class 10 & 0.43 & 0.54 & 0.50 \\
    \hline
    Class 11 & 1.00 & 1.00 & 0.67 \\
    \hline
    Class 12 & 0.46 & 0.45 & 0.30 \\
    \hline
    Class 13 & 0.62 & 0.73 & 0.00 \\
    \hline
    Class 14 & 0.00 & 0.00 & 0.00 \\
    \hline
    Class 15 & 0.53 & 0.52 & 0.57 \\
    \hline
    Class 16 & 0.22 & 0.21 & 0.17 \\
    \hline
    Class 17 & 0.52 & 0.49 & 0.26 \\
    \hline
    Class 18 & 0.91 & 0.91 & 0.91 \\
    \hline
    Class 19 & 0.73 & 0.73 & 0.69 \\
    \hline
    Class 20 & 0.34 & 0.38 & 0.42 \\
    \hline
    Class 21 & 0.67 & 0.62 & 0.50 \\
    \hline
  \end{tabular}
\end{table}

\end{document}